\documentclass{article} 
\usepackage{iclr2027_conference,times}

\usepackage{amsmath,amsfonts,bm}

\def\eqref#1{equation~\ref{#1}}

\def\1{\bm{1}}

\DeclareMathAlphabet{\mathsfit}{\encodingdefault}{\sfdefault}{m}{sl}
\SetMathAlphabet{\mathsfit}{bold}{\encodingdefault}{\sfdefault}{bx}{n}

\usepackage{hyperref}
\usepackage{url}

\usepackage{graphicx}      
\usepackage{booktabs}      
\usepackage{array,multirow} 
\usepackage{pifont}        
\usepackage{enumitem}      
\newcommand{\cmark}{\ding{51}}

\usepackage[most]{tcolorbox}
\newcounter{prompt}
\newtcblisting{promptbox}[1]{%
  enhanced, listing only, title={#1},
  colback=gray!4, colframe=gray!60!black, fonttitle=\bfseries\small,
  boxrule=0.5pt, arc=1pt, left=4pt, right=4pt, top=2pt, bottom=2pt,
  listing options={basicstyle=\ttfamily\scriptsize, breaklines=true,
    breakatwhitespace=true, columns=fullflexible, keepspaces=true,
    literate={—}{{\textrm{\textemdash}}}1}}

\definecolor{actWatch}{HTML}{3D6DB5}
\definecolor{actRemember}{HTML}{3E8E6E}
\definecolor{actRevisit}{HTML}{C07A2C}
\definecolor{actAnswer}{HTML}{5F6368}
\definecolor{memCard}{HTML}{FFF8DF}
\definecolor{memEdge}{HTML}{B9AE7C}
\definecolor{goldTint}{HTML}{D9ECE3}
\newcommand{\act}[2]{\tcbox[on line, colback=#1, colframe=#1, coltext=white, boxsep=0pt,
  left=3pt, right=3pt, top=1.5pt, bottom=1.5pt, arc=2pt, fontupper=\scriptsize\bfseries]{#2}}
\definecolor{callInk}{HTML}{202124}
\definecolor{resBack}{HTML}{F4F5F7}
\newcommand{\call}[1]{\leavevmode\hangindent=1.1em\hangafter=1{\color{callInk}\ttfamily\footnotesize\textcolor{actAnswer}{\scriptsize$\triangleright$}\ #1}}
\newtcolorbox{res}[1]{enhanced, colback=resBack, frame hidden, boxrule=0pt, arc=0pt,
  borderline west={1.6pt}{0pt}{#1}, left=5pt, right=4pt, top=2pt, bottom=2pt,
  before skip=1.5pt, after skip=0pt, fontupper=\footnotesize\color{black!80}}
\newcommand{\trajframe}[3][0.18]{\begin{tabular}[t]{@{}c@{}}{\setlength{\fboxsep}{0pt}\fcolorbox{black!25}{white}{\includegraphics[width=#1\linewidth]{figure/traj/#2}}}\\[-1pt]\scriptsize\textcolor{actAnswer}{#3}\end{tabular}}
\newtcolorbox{memcard}{colback=memCard, colframe=memEdge, boxrule=0.4pt, arc=2pt,
  left=3pt, right=3pt, top=1pt, bottom=1pt, before skip=3pt, after skip=2pt, fontupper=\scriptsize}
\newcommand{\gold}[1]{\colorbox{goldTint}{#1\,\textcolor{actRemember}{\cmark}}}

\usepackage{color}
\definecolor{citecolor}{RGB}{66,168,235}
\definecolor{linkcolor}{RGB}{255,0,0}

\hypersetup{colorlinks=true,citecolor=citecolor,linkcolor=linkcolor}

\newcommand{\methodname}{Sprout}

\title{Sprout: Building Dynamic Memory While Reasoning for Agentic Video Understanding}

\iclrfinalcopy

\author{%
  Wei Chen$^{1}$, Xuanyu Zheng$^{2}$, Yancheng Long$^{2}$, Haoyang Xu$^{2}$, Kaiyu Jiang$^{2}$, \\
  \textbf{Bin Wen$^{2}$, Tingting Gao$^{2}$, Han Li$^{2}$, Long Chen$^{1,\dagger}$} \\[2pt]
  $^{1}$HKUST \qquad $^{2}$Kling AI \\
  \texttt{wchendb@connect.ust.hk}, \texttt{longchen@ust.hk}
}

\begin{document}

\maketitle
\lhead{Preprint}
{\renewcommand{\thefootnote}{\fnsymbol{footnote}}\footnotetext[2]{Corresponding author.}}
\vspace{-0.15in}
{\centering\small\bfseries \href{https://github.com/HKUST-LongGroup/Sprout}{\texttt{https://github.com/HKUST-LongGroup/Sprout}}\par}
\vspace{0.1in}

\begin{abstract}
Long video understanding relies on video memory to overcome the context limits of multimodal large language models.
Existing methods follow a build-then-reasoning pipeline: memory is built \emph{offline} for the entire video, then reasoned over as a \emph{static} source.
In practice a long video is shared by several questions, and this pipeline is costly at both ends: with few questions, building memory for the whole video costs far more than answering them; with many questions, the memory is never updated, so what is learned while answering questions is lost to the next question.
To alleviate these, we introduce \textbf{\methodname{}}, an agentic framework that builds memory \emph{while} reasoning: a temporal tree that sprouts detailed nodes as questions are answered.
The agent \emph{watches} the video segment by segment at a low frame rate, stopping when the current question can be answered, \emph{remembers} each segment as a coarse node of the tree, and \emph{revisits} key intervals at a higher frame rate to refine the tree with the recovered details.
Once a segment is recorded as text, its video input is removed from the context history, while the original video remains reachable through the video tools.
The memory tree and prior question--answer records persist across questions, so the memory is online and dynamic: built from the first question onward and updated by every question thereafter.
We find that replacing accumulated video inputs with textual memory substantially reduces context usage while maintaining accuracy, with slight improvements in some settings.
Across benchmarks on three models, \methodname{} achieves competitive or improved accuracy relative to representative offline memory methods, with no upfront construction stage and lower context cost per question.
\end{abstract}

\section{Introduction}
\label{sec:intro}

Advances in multimodal large language models (MLLMs)~\citep{qwen2026qwen38,gemini2026gemini38flash,keye2026keyevl2} have extended video understanding from short clips to long recordings spanning hours, such as lectures, films, and days of egocentric activity~\citep{fu2025videomme,wang2025lvbench,yang2025egolife}.
Answering questions about such videos requires locating brief moments scattered across time and connecting the evidence among them.
As duration grows, however, feeding the video directly into the model becomes infeasible: the number of visual tokens scales linearly with length and quickly exceeds what a context window can hold.
Sampling more sparsely fits the budget but discards the brief moments that questions often hinge on, while sampling more densely preserves them at a cost the context cannot absorb.
Long video understanding therefore relies on \emph{video memory}: a representation of the video maintained outside the context, from which question-relevant content is retrieved when needed.

\begin{figure}[t]
    \centering
    \includegraphics[width=\linewidth]{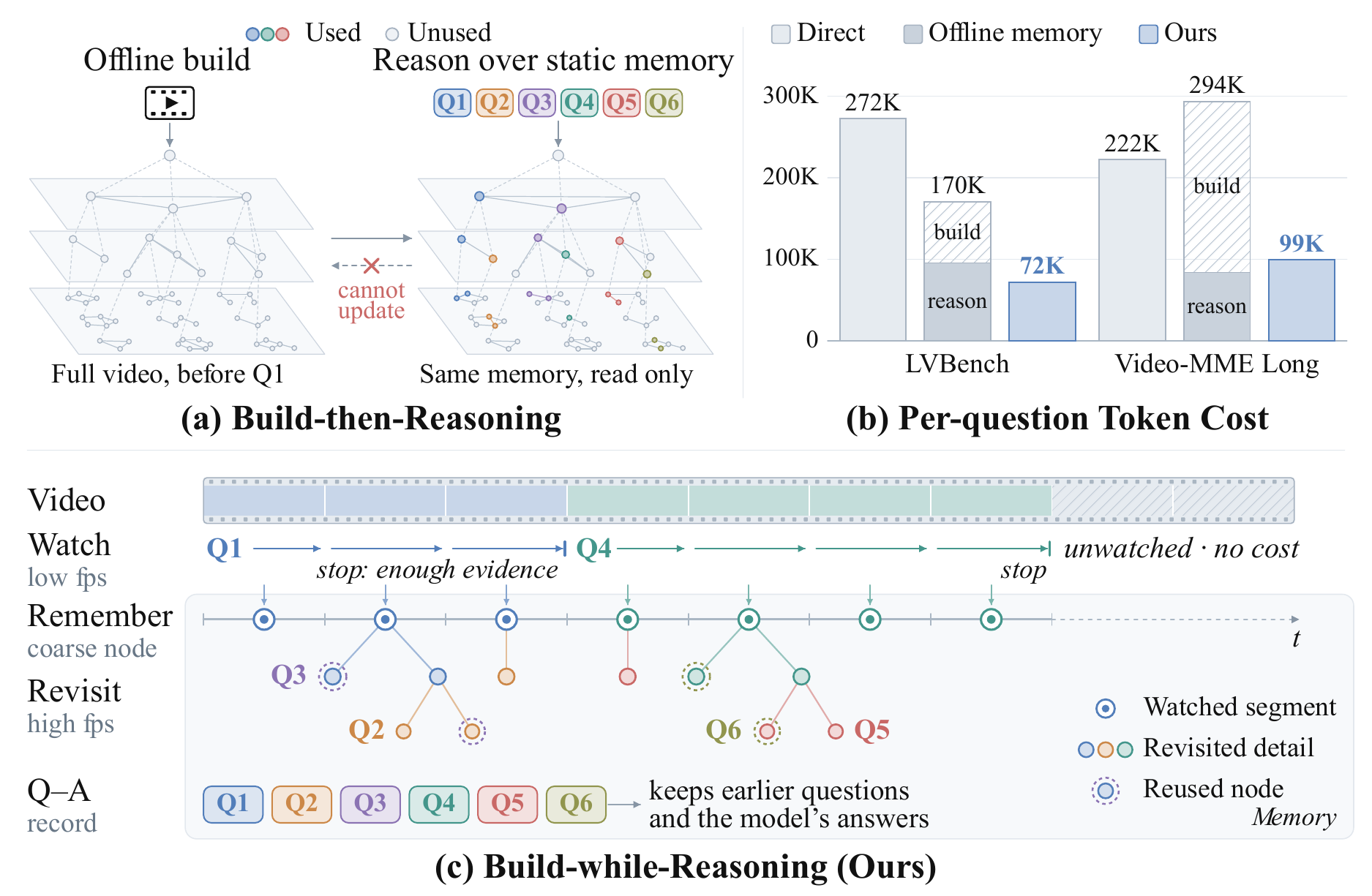}
    \vspace{-2.5em}
    \caption{\textbf{(a)} Offline memory is first built for the whole video before the first question; Q1--Q6 then reason over the same memory without updating it. Nodes used by a question take that question's color, and the rest are never touched.
    \textbf{(b)} Per-question tokens with Gemini 3.8 Flash as the model. \emph{Direct} feeds the video and question directly to the model.  \emph{Offline memory} is Qwen-MM-Plugins: its construction cost, amortized over the benchmark's questions. Our method has no construction stage.
    \textbf{(c)} \methodname{} builds memory while reasoning. The agent \emph{watches} the video segment by segment at a low frame rate and stops once the question has enough evidence, so the tail of the video is never watched and costs nothing. Each watched segment is \emph{remembered} as a coarse textual node written into memory; \emph{revisits} at a higher frame rate add child nodes beneath the intervals a question needs. }
    \label{fig:intro_memory}
\end{figure}

Existing methods~\citep{qwen2026mmplugins,choi2026merit} construct video memory through a \emph{build-then-reasoning} pipeline (Figure~\ref{fig:intro_memory}a): they first build memory \emph{offline}, then reason over the resulting \emph{static} memory with an agentic framework. 
In the construction stage, the model processes the entire video without knowing future questions and stores it as persistent memory; In the reasoning stage, it retrieves relevant evidence from that memory to answer a given question.
Prior work within this pipeline mainly differs in how the offline memory is organized. Hierarchical memories~\citep{yang2025egolife,tian2025egor1} recursively summarize short segments into multi-scale representations, enabling top-down retrieval from coarse summaries to fine segments. Graph-based memories~\citep{long2026m3agent,yeo2026worldmm,chen2026memdreamer,qwen2026mmplugins} model relations among events, entities, and time, answering questions through navigation and traversal over the resulting structure. Flat episodic memories~\citep{choi2026merit} keep construction minimal by indexing clip captions with multiple retrieval keys and deferring semantic composition to inference time. Despite these differences, all share two properties. First, memory is built offline over the entire video before answering, requiring costly conversion of fine-grained details into memory. Second, memory is static: it is only read during inference and not updated, so the model’s observations and conclusions cannot persist across questions.
In practice, a long video may be used to answer anywhere from a single question to a dozen or more. This leaves offline, static memory limited at both extremes: 1) upfront construction is wasteful when few questions are asked, since it must record every segment regardless of what is eventually asked; 2) a non-updatable memory prevents useful information from accumulating when many questions are asked, forcing each question to recover the same evidence from scratch.

To address these limitations, we propose a new framework whose memory is built \emph{while} reasoning rather than before it, following two principles.
First, the memory is built online from what the model has watched: rather than recording every segment in advance, it grows as the model watches for each question, so no cost is paid for video that no question needs.
Second, the memory is refined from coarse to fine: it starts as brief summaries of what happens in each segment, and whenever the model looks closer at a segment to answer a question, it adds the details it finds.

Based on the build-while-reasoning framework, we introduce \textbf{\methodname{}}, an agentic framework whose memory is a temporal tree that sprouts new nodes as questions are answered (Figure~\ref{fig:intro_memory}c), built through three actions: \emph{watch}, \emph{remember}, and \emph{revisit}.
\emph{Watch.} Given a question, the agent streams the video segment by segment at a low frame rate, starting from the beginning of videos or where the memory left off, and stops once it judges that the segments the question needs have been found.
\emph{Remember.} As each segment is watched, the agent writes it into memory as a coarse node with a time interval and an event summary, forming the first level of a temporal tree.
Once a segment is recorded, its video input is removed from the context history, allowing the agent to reason without that video chunk and greatly reducing context pressure. 
\emph{Revisit.} When a retrieved node identifies a relevant video interval but lacks the needed detail, the agent revisits it at a higher frame rate, saves the recovered details as child nodes, and may search memory for related earlier information. The revisited frames are released from the context history in the same way.

Together, the three actions follow the two principles above.
Watching and remembering build the memory online from the video the model has actually watched, and keep text rather than video in the context.
Revisiting refines coarse nodes into finer ones while the model reasons.
As a result, our memory is \emph{online} and \emph{dynamic}: it is built online after the first question arrives, even when the full video is available in advance, and it keeps growing as each question adds nodes, refines existing ones, and leaves its answer behind.

We evaluate \methodname{} with three models (GPT-5~\citep{openai2025gpt5}, Qwen3.8-Max, and Gemini 3.8 Flash) on LVBench~\citep{wang2025lvbench}, Video-MME Long~\citep{fu2025videomme}, and Video-Holmes~\citep{cheng2026videoholmes}.
Its accuracy is competitive with or better than representative offline memory methods, while it pays no construction cost before the first question.
It also uses fewer tokens (Figure~\ref{fig:intro_memory}b): with Gemini 3.8 Flash, it spends 71.7K and 99.3K tokens per question on LVBench and Video-MME Long, 74\% and 55\% fewer than direct inference.

In summary, our contributions are threefold:
\begin{itemize}[left=0.5em, labelsep=0.5em]
    \item We identify two limitations of the build-then-reasoning pipeline: offline construction is wasteful for few-question videos, and static memory cannot accumulate insights from earlier reasoning.
    \item We propose \methodname{}, an agentic framework that builds memory while reasoning through three actions: it watches the video at a low frame rate, remembers each segment as a coarse node of a temporal tree, and revisits key intervals at a higher frame rate to add finer nodes. 
    \item We further find that removing video frames from the context once they are recorded as text keeps the context small without reducing accuracy. Experiments with three models on benchmarks show that \methodname{} is more effective than an offline memory method.
\end{itemize}
\section{Related Work}
\label{sec:related}

\noindent\textbf{Agentic Multimodal Reasoning. }
Rather than answering from what is already in the context, agentic multimodal methods call external tools during reasoning to gather the evidence they need.
MM-REACT routes a language model to vision experts through a textual interface~\citep{yang2023mmreact}, while ViperGPT composes visual and language modules into executable programs~\citep{suris2023vipergpt}.
For video, VideoAgent iteratively gathers relevant visual information under the control of an LLM~\citep{wang2024videoagent}.
DrVideo converts video into a coarse textual document and augments selected entries with additional visual information as its agent searches for missing evidence~\citep{ma2025drvideo}.
More recently, VideoSeek provides overview, skim, and focus tools, allowing an agent to adapt its inspection granularity while reasoning over accumulated conversation history~\citep{lin2026videoseek}.
VideoSearcher instead trains the tool-invocation policy itself with reinforcement learning, unifying temporal localization, spatial focusing, and external multimodal search for open-world video questions~\citep{gao2026videosearcher}.
These methods show that tool use during reasoning can improve multimodal performance.
More broadly, reliable multimodal reasoning rests on grounding outputs in explicit visual evidence: models comprehend relations among objects~\citep{li2026relationr1}, keep interleaved image--text content coherent~\citep{chen2025comm}, suppress hallucinations unsupported by the image~\citep{chen2025dcd}, and ground reward judgments in fine-grained, region-level visual evidence~\citep{long2026spatialreward,yang2026jrm,yang2026spatialflowgrpo}.
\methodname{} connects this tool use to persistent memory management: each inspection updates a temporal node, its video inputs can then leave the context history, and the recorded information remains available to subsequent questions.

\noindent\textbf{Efficient Long Video Understanding. }
Long video understanding requires balancing temporal coverage, visual detail, and input cost, as reflected in Video-MME, LVBench, and EgoLife~\citep{fu2025videomme,wang2025lvbench,yang2025egolife}.
One direction expands the amount of visual information a model can process: LongVILA combines context extension, long-video instruction tuning, and sequence parallelism to support longer video inputs~\citep{chen2025longvila}, and Keye-VL-2.0 adapts sparse attention to a multimodal backbone to process hour-level videos within a 256K-token context~\citep{keye2026keyevl2}.
Another direction reduces the representation size: LongVU adaptively compresses spatial and temporal visual tokens to fit long videos within a limited context~\citep{shen2025longvu}.
At inference time, VideoTree allocates representation detail according to query relevance through adaptive expansion of a video tree~\citep{wang2025videotree}.
Together, these approaches improve the trade-off between coverage and detail through model capacity, compression, or evidence selection.
Our focus is how observation density and retained context evolve over an answering session.
\methodname{} first builds coarse temporal nodes from low-frame-rate observations, then revisits relevant intervals at higher frame rates and writes the recovered details back to those nodes.
Textual memory carries these observations into later reasoning steps, allowing the corresponding visual inputs to be released from the context history while the source video remains accessible.

\noindent\textbf{Video Memory and Retrieval. }
Long-video methods keep a memory outside the context and answer by retrieving from it.
This memory follows a build-then-reasoning pipeline: it is built before any question, in one pass or incrementally as the video arrives, and it stays \emph{static} while questions are answered.
Streaming methods such as MA-LMM and M3-Agent are the incremental case: their memory grows as frames arrive, but it is still formed without questions and only read at answer time~\citep{he2024malmm,long2026m3agent}.
Within this pipeline, methods differ in how the memory is organized.
Hierarchical memories summarize segments at multiple temporal scales and answer by top-down retrieval~\citep{yang2025egolife,tian2025egor1}.
Graph-based memories link events, entities, and time, then answer by search or traversal: WorldMM, MemDreamer, and Qwen-MM-Plugins retrieve over multi-scale or hierarchical graphs~\citep{yeo2026worldmm,chen2026memdreamer,qwen2026mmplugins}; Vgent and EGAgent add verification or entity search on the graph~\citep{shen2025vgent,rege2026egagent}; HippoMM consolidates short-term observations into longer-term memory~\citep{lin2026hippomm}.
MERIT keeps the memory flat, indexing clip captions under multiple keys and assembling evidence at inference time~\citep{choi2026merit}.
Caption-based methods such as LLoVi, Video-RAG, and iRAG let the question decide which text is fetched or refined~\citep{zhang2024llovi,luo2025videorag,arefeen2024irag}.
\methodname{} builds the memory while reasoning, adding and refining nodes only where questions lead and keeping those nodes and earlier answers for later questions.

\section{Method}
\label{sec:method}

\begin{figure}[t]
    \centering
    \includegraphics[width=\linewidth]{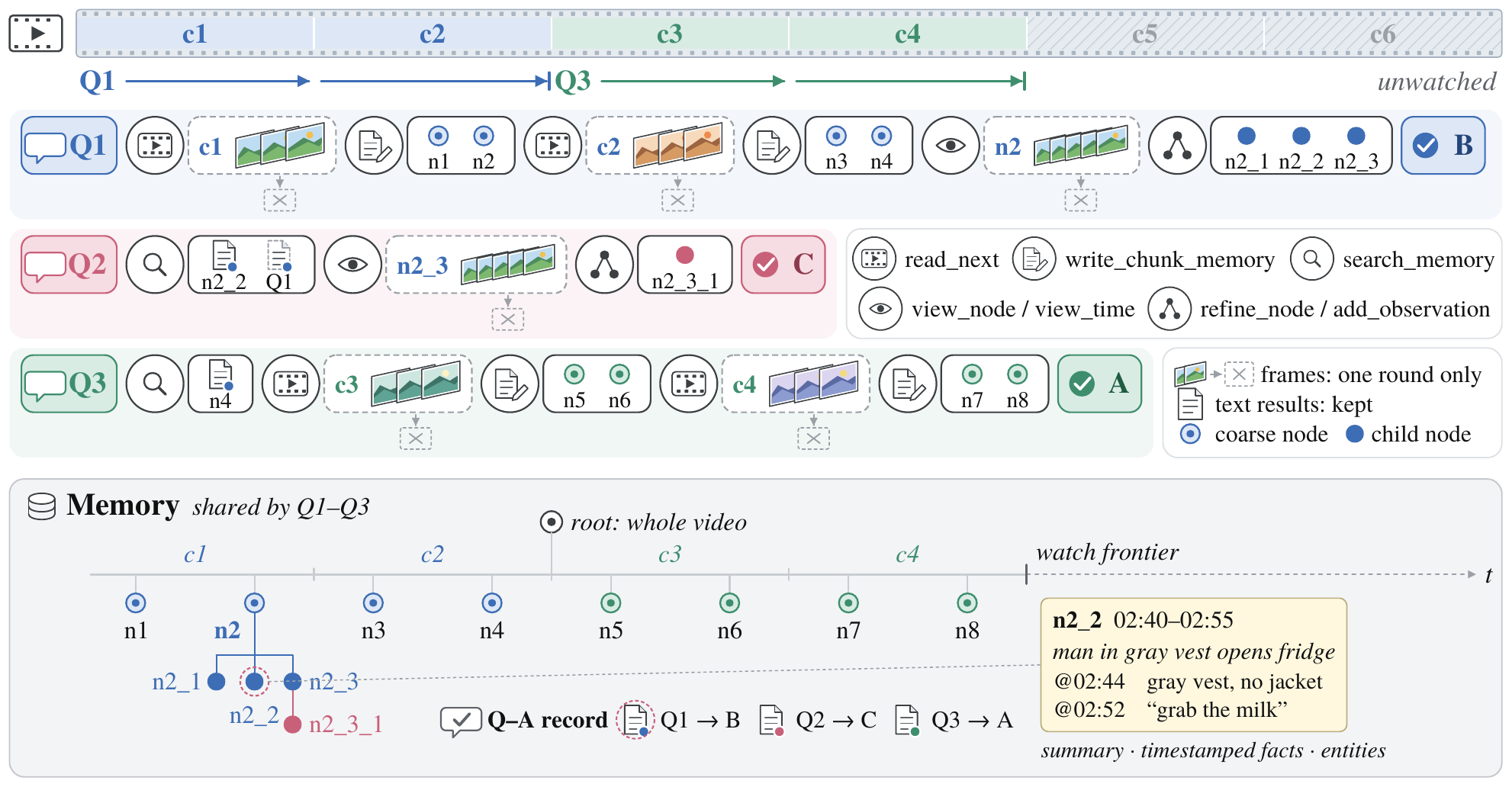}
    \vspace{-1.7em}
    \caption{\textbf{Pipeline of \methodname{}.}
    Three questions about one video are answered in order, building one memory.
    Each row shows the rounds of one question: tool calls (circles) and their results (boxes).
    Video frames stay in the context for one round only; text results are kept.
    \textbf{Q1} watches the first two chunks, remembers them as coarse nodes, and revisits $n_2$ at a higher frame rate to add finer child nodes.
    \textbf{Q2} reuses the memory: search returns a node and Q1's answer, and a short revisit adds one more detail, without watching new video.
    \textbf{Q3} finds that the memory does not cover what it asks, so it watches two more chunks.
    The last two chunks are never watched.
    The memory tree (bottom) is drawn on the same time axis as the video; node colors show which question added each node.}
    \label{fig:method}
\end{figure}

In this section, we present \methodname{}, an agentic framework that builds video memory while reasoning.
Unlike build-then-reasoning methods, which construct memory over the entire video before any question arrives, \methodname{} starts from an empty memory and builds it only while answering questions.
Given a video $V$ of duration $T$ and questions $q_1,\ldots,q_K$ about it, the model answers each question over multiple rounds, calling tools (Appendix~\ref{app:tools}) that perform three actions: \emph{watch} attaches the next unwatched chunk of the video at a low frame rate, \emph{remember} writes it into memory as text, and \emph{revisit} watches a recorded interval again at a higher frame rate to add details.
The memory is shared by all questions about the video, so later questions can reuse what earlier ones have built.
Figure~\ref{fig:method} illustrates this process on three questions.
\subsection{Dynamic Temporal Memory}
\label{sec:method:memory}
The memory has three parts: a temporal tree of observations, a watch record of how far the video has been watched, and a question history of earlier questions and their answers.

\noindent\textbf{Temporal tree.}
The tree is organized by time.
Its root spans the whole video $[0,T]$.
Every other node covers an interval $[s_i,e_i]\subseteq[0,T]$ and stores an event summary, observed facts, and key entities for that interval; the children of a node split its interval into shorter pieces in temporal order.
Because each node keeps its interval, the video behind any node can be watched again.
The tree grows with the three actions: watching and remembering add coarse nodes under the root in temporal order, and revisiting adds children under a node where a question needs more detail.
At any moment, the tree is therefore deep only where questions have looked closely, and it ends where the unwatched video begins.
Nodes store only what was observed in the video; questions and answers are kept in the question history, so that the nodes remain useful to unrelated later questions.

\noindent\textbf{Access.}
The model does not see the tree directly.
Its context holds the current question, the watch record, and a compact list of the question history; the content of the tree is reached only through search (Section~\ref{sec:method:revisit}).
All writes go through tools that check the new nodes against the current tree, so that at every level the tree stays in chronological order within the watched video.

\subsection{Watch: Coarse Temporal Exploration}
\label{sec:method:watch}
The agent watches new video with \texttt{read\_next}, either when the memory is empty, as at the first question, or when the memory does not yet cover what the question asks about.
The tool takes no arguments: the runtime selects the next unwatched chunk from the watch record and attaches its frames at a low rate $f_{\mathrm{c}}$, together with the speaker-labelled transcript of the chunk when audio is available.
At this rate, the agent learns what happens and when at a small visual cost.

How far to watch is decided by the current question.
After each chunk is remembered (Section~\ref{sec:method:remember}), the agent has three options: watch the next chunk, revisit an interval that looks relevant, or answer.
Watching can therefore stop before the end of the video, and dense revisits can already happen during the first question.
Everything watched stays in memory for later questions.

\subsection{Remember: Recording Observations and Releasing Frames}
\label{sec:method:remember}
Each watched chunk is written into memory right away.
After \texttt{read\_next}, the chunk is \emph{pending}: the agent must store it with \texttt{write\_chunk\_memory} in the next round before it can watch another chunk.
The call passes a list of segments, each with an interval, a summary, and a set of facts.
The runtime appends them as coarse nodes under the root and advances the watch record to the end of the chunk.
The agent is instructed to write about 4--8 segments per chunk with 5--10 objective facts each, and to record only what is in the video, never the question, its options, or its reasoning.

\noindent\textbf{Releasing frames.}
Once the model has responded to a chunk, its frames are removed from the context history before the next model call; the textual tool results and the model's responses are kept.
Frames therefore never pile up in the context: each chunk costs one call with frames and one text-only call to store it.
Frames from revisits are removed in the same way.

To see the saving, consider $t$ observations with $v_1,\ldots,v_t$ visual tokens and a textual context of length $B_t$.
Keeping all frames and releasing them after use give context lengths
{\setlength{\abovedisplayskip}{0.35em}
\setlength{\abovedisplayshortskip}{0.35em}
\setlength{\belowdisplayskip}{0.35em}
\setlength{\belowdisplayshortskip}{0.35em}
\begin{equation}
    L_t^{\mathrm{retain}} = B_t + \sum_{j=1}^{t}v_j,
    \qquad
    L_t^{\mathrm{release}} = B_t + v_t.
    \label{eq:visual_context}
\end{equation}
}
Each observation is still paid for once, but only the latest one stays in the context.

\subsection{Revisit: Retrieval, Dense Inspection, and Write-Back}
\label{sec:method:revisit}
Revisiting lets the agent return to video it has already watched.
It has three steps: retrieve relevant nodes from the memory, watch their intervals at a higher frame rate, and write new details back.

\noindent\textbf{Retrieval.}
Since the model does not see the tree, it searches the memory with \texttt{search\_memory}, which returns text only.
A query is a set of literal, case-insensitive alternatives (e.g., \texttt{"vest | waistcoat"}), a time window, or both; a window alone returns everything stored in that span (Appendix~\ref{app:search} compares this literal matching with embedding retrieval).
The result lists the matching facts with their timestamps under their nodes, followed by entries of the question history whose question mentions the query, so earlier answers come back together with the observations behind them.
The agent is instructed to search before answering detail questions, and the number of searches per question is capped.

\noindent\textbf{Dense inspection.}
When a retrieved node points to the right event but lacks the detail the question needs, the agent watches that part of the video again.
\texttt{view\_node} attaches the whole interval of a node, and \texttt{view\_time} attaches an explicit interval inside the watched video, which the runtime clamps to the node that contains it.
Both tools take a \emph{focus} describing what to look for and a frame rate $f_{\mathrm{r}}>f_{\mathrm{c}}$ chosen by the agent, so dense frames are spent only on the chosen interval.
This can reveal brief events or intermediate actions missed at the low rate.

\noindent\textbf{Write-back.}
Each inspection tool is paired with a write-back tool that accepts only the interval just viewed.
\texttt{refine\_node} splits the node just viewed with \texttt{view\_node} into contiguous children in temporal order that cover its interval.
This is allowed only for a leaf, so an internal node can serve as evidence but its subtree is never rewritten.
\texttt{add\_observation} stores a summary and facts for the interval just viewed with \texttt{view\_time} as a child of the node.
In both cases, the parent keeps its description, so coarse context and new detail are kept side by side, and the revisited frames are removed as in Section~\ref{sec:method:remember}.
If the evidence is still not enough, the agent can search again, revisit other nodes, or watch more video.
A question ends when the model calls \texttt{answer} or reaches the maximum number of rounds, in which case it must answer with the evidence it has.

\subsection{Accumulation Across Questions}
\label{sec:method:accumulation}
After a question is answered, the tree and the watch record are kept for the next question.
The \texttt{answer} call also appends the question, its options, and the predicted answer to the question history; later questions see these entries in compact form and can retrieve them with \texttt{search\_memory}.
A new question therefore starts from everything earlier questions have built: it can reuse existing nodes, refine them, or extend the tree into unwatched video.
Since all watching and writing happens while questions are answered, the memory covers only what the questions have needed, and each question pays only for the video and detail it adds.

\section{Experiments}
\label{sec:exp}

\subsection{Experimental Setup}
\label{sec:exp:setup}

\noindent\textbf{Benchmarks and metric.}
We evaluate on three complementary benchmarks: LVBench~\citep{wang2025lvbench} contains 1,549 questions over 103 hour-scale videos; the Long subset of Video-MME~\citep{fu2025videomme} contains 900 questions over 300 videos lasting 30--60 minutes; and Video-Holmes~\citep{cheng2026videoholmes} contains 1,837 questions over 270 short suspense videos, emphasizing reasoning across scattered visual clues.
We also evaluate on EgoLifeQA~\citep{yang2025egolife}, using the 500 questions of one participant (Jake) over a 44.3-hour recording spanning seven days.
With hundreds of questions about the same recording, it tests how well the memory is reused.
Table~\ref{tab:main} reports the first three benchmarks and Table~\ref{tab:egolifeqa} reports EgoLifeQA.

\noindent\textbf{Implementation.}
We run \methodname{} with three models: GPT-5, Qwen3.8-Max, and Gemini 3.8 Flash.
The memory starts empty for each video and is kept across its questions.
Questions arrive one at a time; the agent sees its own earlier answers, but not the ground truth or future questions.
On LVBench and Video-MME Long, \texttt{read\_next} attaches 10-minute chunks at $f_{\mathrm{c}}=0.1$~fps; GPT-5 uses 5-minute chunks at the same rate to fit the image limit of its endpoint.
On Video-Holmes, chunks are 30 seconds long at $f_{\mathrm{c}}=1$~fps.
Revisits choose $f_{\mathrm{r}}\in\{1,2,4,8\}$~fps and attach at most 768, 48, and 2,048 frames per view for Qwen3.8-Max, GPT-5, and Gemini, respectively.
For comparison, Direct gives Qwen3.8-Max up to 768 sampled frames on Video-MME Long and 1~fps on Video-Holmes, and gives Gemini up to 3,072 frames.
Each question may use at most 32 \emph{rounds}.
An ASR transcript is attached to each chunk on Video-MME Long, but not on LVBench or Video-Holmes.
As described in Section~\ref{sec:method:memory}, the model does not see the tree: its context shows only the watch record and the question history, and the tree is reached through \texttt{search\_memory}.
We call this setting the \emph{progress exposure}.
For Qwen-MM-Plugins, the memory is built once by Qwen3.7-Plus and shared by all three models.

\subsection{Main Results}
\label{sec:exp:main}
\begin{table}[t]
\caption{\textbf{Accuracy (\%) on three video benchmarks.}
Rows are grouped into direct inference, agentic and memory methods, and \methodname{}; Video-MME(L) is the Long subset of Video-MME.}
\label{tab:main}
\centering
\footnotesize
\setlength{\tabcolsep}{3pt}
\begin{tabular*}{\linewidth}{@{\extracolsep{\fill}}llrrr@{}}
\toprule
Method & Model & LVBench & Video-MME(L) & Video-Holmes \\
\midrule
Direct & GPT-5 & 60.4 & 74.3 & 44.1 \\
Direct & Qwen3.8-Max & 81.8 & 80.4 & 67.3 \\
Direct & Gemini 3.8 Flash & 87.1 & 90.1 & 72.0 \\
\midrule
WorldMM~\citep{yeo2026worldmm} & GPT-5 & 61.9 & 76.6 & -- \\
MERIT~\citep{choi2026merit} & GPT-5 & 71.8 & 77.7 & -- \\
VideoSeek~\citep{lin2026videoseek} & GPT-5 & 68.4 & 81.2 & 47.3 \\
Qwen-MM-Plugins & GPT-5 & 73.3 & 79.2 & 48.4 \\
Qwen-MM-Plugins & Qwen3.8-Max & 85.6 & 85.8 & 68.5 \\
Qwen-MM-Plugins & Gemini 3.8 Flash & 85.5 & 89.4 & 69.2 \\
\midrule
\methodname{} (ours) & GPT-5 & 73.4 & 80.4 & 56.7 \\
\methodname{} (ours) & Qwen3.8-Max & 86.9 & 86.0 & 71.6 \\
\methodname{} (ours) & Gemini 3.8 Flash & 88.2 & 91.2 & 76.8 \\
\bottomrule
\end{tabular*}
\end{table}

\begin{table}[t]
\caption{\textbf{Accuracy (\%) on EgoLifeQA.}
Ent., Event, Habit, Rel., and Task denote the EntityLog, EventRecall, HabitInsight, RelationMap, and TaskMaster categories; Overall is the accuracy over all questions, not the mean of the categories.}
\label{tab:egolifeqa}
\centering
\footnotesize
\setlength{\tabcolsep}{3pt}
\begin{tabular*}{\linewidth}{@{\extracolsep{\fill}}llrrrrrr@{}}
\toprule
Method & Model & Ent. & Event & Habit & Rel. & Task & Overall \\
\midrule
Direct & Gemini 2.5 Pro & 43.2 & 40.5 & 41.0 & 55.2 & 52.4 & 46.4 \\
Ego-R1~\citep{tian2025egor1} & 3B & 51.2 & 53.2 & 63.9 & 50.4 & 50.8 & 53.0 \\
M3-Agent~\citep{long2026m3agent} & 7B & 44.4 & 54.8 & 62.3 & 56.8 & 54.0 & 53.5 \\
WorldMM~\citep{yeo2026worldmm} & Qwen3-VL-8B & 49.6 & 56.4 & 63.9 & 58.4 & 58.7 & 56.4 \\
MERIT~\citep{choi2026merit} & Gemini 2.5 Pro & 60.8 & 61.1 & 65.6 & 65.6 & 73.0 & 64.2 \\
Qwen-MM-Plugins & Qwen3.8-Max & 72.0 & 77.0 & 80.3 & 72.8 & 82.5 & 75.8 \\
Qwen-MM-Plugins & Gemini 3.8 Flash & 76.0 & 77.0 & 77.0 & 75.2 & 87.3 & 77.6 \\
\midrule
\methodname{} (ours) & Qwen3.8-Max & 79.2 & 77.0 & 80.3 & 84.0 & 84.1 & 80.6 \\
\methodname{} (ours) & Gemini 3.8 Flash & 76.0 & 75.4 & 85.2 & 86.4 & 81.0 & 80.2 \\
\bottomrule
\end{tabular*}
\end{table}

\noindent\textbf{Long videos.}
Table~\ref{tab:main} reports results on LVBench and Video-MME Long.
Memory built offline does not always help: with Gemini 3.8 Flash, Qwen-MM-Plugins even falls below Direct.
In contrast, \methodname{} matches or exceeds both with every model, although it builds no memory before the first question.
With Gemini 3.8 Flash, it achieves new state-of-the-art results on both benchmarks.
The gain over Direct is largest with GPT-5 (13.0 points on LVBench) and small with Gemini 3.8 Flash, so the memory helps most when the model cannot use a long context well.

\noindent\textbf{Short videos.}
The advantage is larger on Video-Holmes, where \methodname{} achieves the best results with all three models.
Offline memory again falls below Direct with Gemini 3.8 Flash, while \methodname{} exceeds Qwen-MM-Plugins by up to 8.3 points.
These results show that \methodname{} is a general method that applies to both long and short videos.

\noindent\textbf{Day-long video.}
Table~\ref{tab:egolifeqa} reports results on EgoLifeQA, whose videos are much longer than those above.
With both Qwen3.8-Max and Gemini 3.8 Flash, \methodname{} outperforms Qwen-MM-Plugins overall, and its Qwen3.8-Max run achieves the best result in the table.
This shows that \methodname{} also performs well on ultra-long videos.

\subsection{Releasing Frames and Context Window}
\label{sec:exp:vishist}
\begin{figure}[t]
    \centering
    \includegraphics[width=\linewidth]{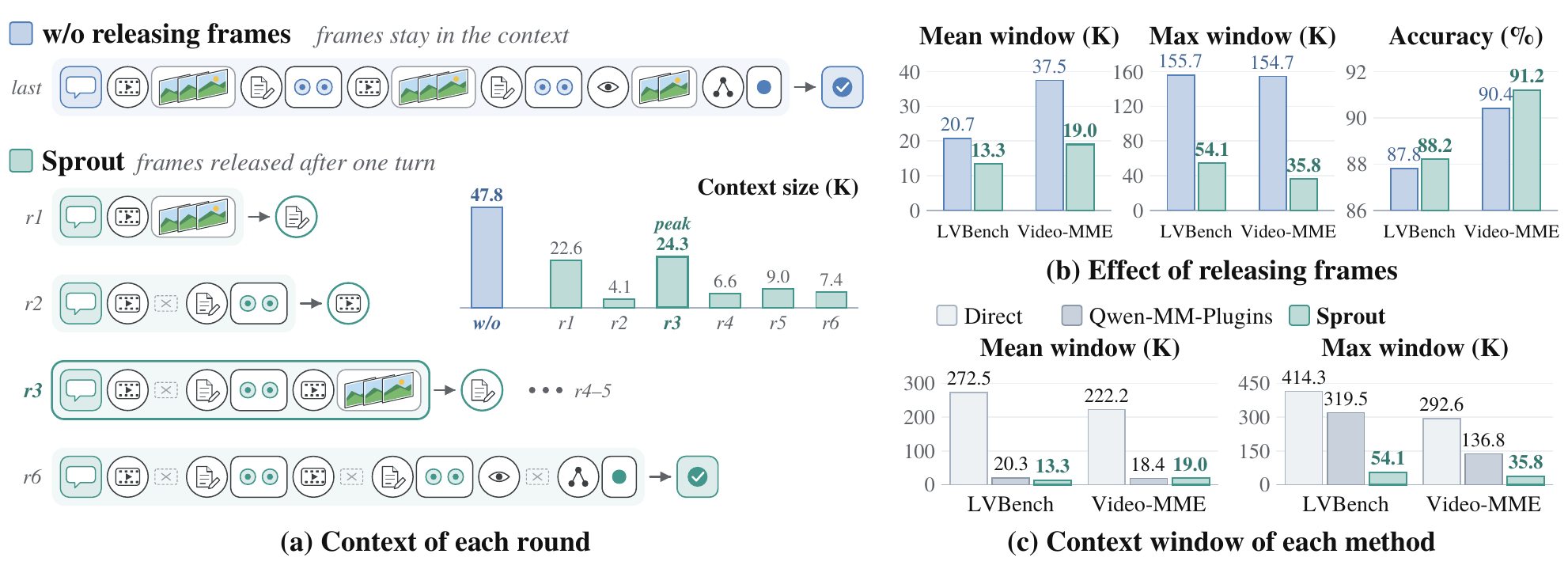}
    \vspace{-1.7em}
    \caption{\textbf{Effect of releasing frames} with Gemini 3.8 Flash.
    (a) The context of one LVBench question in each round, followed by the model output of that round; the bar chart gives the measured context size of each round.
    Without releasing frames, every frame stays, so the context only grows and its last round reaches 47.8K tokens.
    \methodname{} releases frames once the model has written them as text, so its context rises and falls: the rounds that watch a chunk (r1, r3) have the most tokens, and even the peak (24.3K) is about half of the variant's last round.
    (b) Context window and accuracy of the two variants.
    (c) Context window of Direct, Qwen-MM-Plugins, and \methodname{}.
    The context window is the peak context size over all rounds of a question (the single request for Direct), in thousands of tokens (K); we report its mean and maximum over all questions.}
    \label{fig:frame_removal}
\end{figure}

As described in Section~\ref{sec:method:remember}, \methodname{} releases frames from the context once they have been recorded as text.
We now examine how this affects accuracy and the context window, with Gemini 3.8 Flash as the model.
Since \methodname{} answers a question over multiple rounds, its context size changes from round to round (Figure~\ref{fig:frame_removal}a): a round that carries frames uses more tokens than the rounds after it, so the last round is not necessarily the one with the most tokens.
We therefore take the peak context size over all rounds of a question as its context window, and report the mean and maximum of this value over all questions.

\noindent\textbf{Releasing frames.}
Figure~\ref{fig:frame_removal}b shows that releasing frames does not hurt accuracy.
We compare \methodname{} with a variant \emph{w/o releasing frames}, which keeps every frame in the context until the question ends.
Keeping frames enlarges the context window by up to 4.3$\times$. Yet accuracy does not improve.
The variant revisits less often because old frames remain visible, but what these frames provide is already recorded in memory or can be recovered by a revisit.

\noindent\textbf{Context window.}
Releasing frames also keeps the context small throughout a question, as shown in Figure~\ref{fig:frame_removal}c.
Direct places the whole sampled video in a single request of more than 200K tokens.
Qwen-MM-Plugins keeps all retrieved text and frames across rounds, so its context grows as a question proceeds, and its largest question approaches the size of Direct.
In contrast, \methodname{} watches each video segment only once and then removes it from the context history.
Its mean window is similar to that of Qwen-MM-Plugins, but its largest window is several times smaller and stays below 55K tokens on both benchmarks.
Therefore, recording each observation as text and releasing its frames keeps the context small without losing accuracy.

\subsection{Cost Analysis}
\label{sec:exp:cost}
\begin{table}[t]
\begin{minipage}[t]{0.51\linewidth}
\caption{\textbf{Token cost per question} with Gemini 3.8 Flash, in thousands of tokens (K) on LVBench~/\ Video-MME(L), the Long subset of Video-MME.
\emph{Construction} builds the memory offline, amortized over the 1,549 and 900 questions of the two benchmarks;
\emph{Answering} sums input and output over all rounds of a question;
\emph{Total} is their sum, with the lowest in bold.}
\label{tab:cost}
\vspace{4pt}
\centering
\scriptsize
\setlength{\tabcolsep}{2pt}
\begin{tabular*}{\linewidth}{@{\extracolsep{\fill}}lccc@{}}
\toprule
Method & Construction & Answering & Total \\
\midrule
Direct & -- & 272.5\,/\,222.2 & 272.5\,/\,222.2 \\
Qwen-MM-Plugins & 75.2\,/\,210.0 & 95.3\,/\,83.7 & 170.5\,/\,293.7 \\
\methodname{} (ours) & -- & 71.7\,/\,99.3 & \textbf{71.7}\,/\,\textbf{99.3} \\
\bottomrule
\end{tabular*}
\end{minipage}%
\hfill
\begin{minipage}[t]{0.45\linewidth}
\caption{\textbf{Effect of the question history} with Gemini 3.8 Flash; accuracy (\%), with Video-MME(L) the Long subset of Video-MME.
\textbf{Direct} answers each question independently.
\textbf{Causal} answers the questions of a video in order: the whole video is attached to the first question, and each later question sees all earlier questions, the model's reasoning, and its answers as text.}
\label{tab:causal}
\vspace{4pt}
\centering
\scriptsize
\setlength{\tabcolsep}{2pt}
\begin{tabular*}{\linewidth}{@{\extracolsep{\fill}}lccc@{}}
\toprule
Method & LVBench & Video-MME(L) & Video-Holmes \\
\midrule
Direct & 87.1 & 90.1 & 72.0 \\
Causal & 87.0\,($-$0.1) & 90.9\,(+0.8) & 75.9\,(+3.9) \\
\bottomrule
\end{tabular*}
\end{minipage}%

\end{table}
We next measure the total tokens a question uses over all its rounds, with Gemini 3.8 Flash (Table~\ref{tab:cost}); for Qwen-MM-Plugins, we add its offline construction cost, spread over the questions of each benchmark.
\methodname{} uses 74\% and 55\% fewer tokens than Direct on LVBench and Video-MME Long, since it watches each segment only once at a low frame rate and revisits only the intervals a question needs.
Qwen-MM-Plugins answers with about as many tokens as \methodname{}, but it pays for construction no matter how many questions a video receives: on Video-MME Long, with only 3 questions per video, construction alone costs 210K tokens per question and pushes the total above Direct.
In contrast, \methodname{} builds memory only as questions arrive, so its total cost is the lowest on both benchmarks.

\subsection{Effect of the Question History}
\label{sec:exp:history}
\methodname{} keeps the questions of a video and its own answers in the question history, which later questions can see.
To test whether this history helps on its own, we separate it from the memory tree (Table~\ref{tab:causal}): Direct answers each question independently, while Causal answers the questions of a video in order and shows each later question the earlier questions, reasoning, and answers as text.
Although these earlier answers may be wrong, Causal does not hurt accuracy.
On Video-Holmes, where questions about one video build on the same clues, accuracy improves by 3.9 points; on LVBench and Video-MME Long, where questions are largely unrelated, the differences stay within 1 point.
Earlier answers thus help when questions share content and do no harm otherwise, so keeping the question history is worth its small text cost.

\section{Conclusion}
\label{sec:conclusion}

We presented \methodname{}, an agentic framework that builds video memory while reasoning instead of before it.
The agent watches the video in coarse chunks, remembers each chunk as a node of a temporal tree, and revisits the intervals a question needs at a higher frame rate, writing the recovered details back into the tree.
Frames are released once they are recorded as text, and the tree and the question--answer records persist across questions, so the memory is built from the first question onward and refined by every question thereafter.
Across three benchmarks and three models, \methodname{} matches or exceeds offline memory methods without an upfront construction stage, keeps the context window bounded by a single observation, and spends fewer tokens per question on long videos.
Our experiments also show that releasing frames after use costs no accuracy and that earlier answers help later questions when they share content.
Future work includes richer retrieval over the tree, such as multi-key indexing, and mechanisms for merging or forgetting nodes, which the current memory, like existing offline memories, does not have.


\bibliography{iclr2027_conference}
\bibliographystyle{iclr2027_conference}

\appendix
\clearpage

\section*{Appendix}

\section{Tool Interface}
\label{app:tools}
Table~\ref{tab:tools} lists the tools that implement the three actions of \methodname{} (Section~\ref{sec:method}).

\begin{table}[h]
\caption{Tool interface of \methodname{}.
Frames attached by a tool stay in the context for the current round only; each write-back tool accepts only the chunk or interval watched in the preceding round.}
\label{tab:tools}
\centering
\footnotesize
\setlength{\tabcolsep}{6pt}
\renewcommand{\arraystretch}{1.15}
\begin{tabular}{@{}l >{\raggedright\arraybackslash}p{0.30\linewidth} >{\raggedright\arraybackslash}p{0.54\linewidth}@{}}
\toprule
Action & Tool (arguments) & Role \\
\midrule
Watch
  & \texttt{read\_next} ()
  & Attach the next unwatched chunk at the low frame rate $f_{\mathrm{c}}$. \\
\addlinespace
Remember
  & \texttt{write\_chunk\_memory} (segments)
  & Store the pending chunk as contiguous coarse memory nodes. \\
\addlinespace
\multirow[t]{5}{*}{Revisit}
  & \texttt{search\_memory} (pattern, window)
  & Retrieve facts and prior questions by literal alternatives and/or a time window. \\
  & \texttt{view\_node} (node\_id, focus, fps)
  & Re-attach the whole interval of an existing node at the chosen frame rate $f_{\mathrm{r}}$. \\
  & \texttt{view\_time} (start, end, focus, fps)
  & Attach a short interval inside the watched video at the chosen frame rate $f_{\mathrm{r}}$. \\
  & \texttt{refine\_node} (node\_id, segments)
  & Subdivide the leaf just viewed into contiguous children. \\
  & \texttt{add\_observation} (summary, facts)
  & Attach what was just viewed with \texttt{view\_time} as a timestamped child of its node. \\
\bottomrule
\end{tabular}
\end{table}

\section{Literal Matching versus Embedding Retrieval}
\label{app:search}
\methodname{} and Qwen-MM-Plugins both let the agent search a textual memory, but they match queries to memory in different ways.

\noindent\textbf{How each method searches.}
In \methodname{}, \texttt{search\_memory} uses literal matching.
A query is a list of alternatives separated by ``$|$'', and a fact matches if it contains any alternative as a case-insensitive substring; an optional time window restricts the search to one span.
The tool returns every matching fact in temporal order under its node, up to 40 facts, followed by up to five earlier questions that mention the query.
Qwen-MM-Plugins uses embedding retrieval.
When its memory is built, every node of the graph is encoded by a text embedding model (\texttt{qwen3-vl-embedding}).
At query time, the query is encoded by the same model, nodes are ranked by cosine similarity and by BM25, the two rankings are fused, and the top 10 nodes are returned.

\noindent\textbf{Why literal matching is enough.}
Current agent models are already strong at searching: they can choose keywords, list synonyms in one query, as in \texttt{"vest | waistcoat | tank top"}, and rewrite the query when a search returns nothing.
With such an agent, simple literal matching finds the facts it needs, so \methodname{} does not rely on an extra text embedding model.
This saves both time and tokens.
In Qwen-MM-Plugins, embedding the nodes alone costs 7.5M, 7.9M, and 1.1M tokens of construction on LVBench, Video-MME Long, and Video-Holmes, and every search waits for one more call to the embedding model.
Literal matching instead runs directly on the current memory tree, with no embedding cost and no index to rebuild when new nodes are written.

\noindent\textbf{Comparison.}
To compare the two directly, we replace literal matching in \texttt{search\_memory} with embedding retrieval, using the same text embedding model as Qwen-MM-Plugins, and keep everything else unchanged.
Table~\ref{tab:search} reports the results with Qwen3.8-Max.
The two perform within one point of each other: literal matching is 0.8 and 0.4 points higher on LVBench and Video-MME Long, and embedding retrieval is 0.5 points higher on Video-Holmes.
Literal matching therefore gives similar accuracy without the cost of an embedding model.

\begin{table}[h]
\caption{\textbf{Literal matching versus embedding retrieval} in \texttt{search\_memory}, with Qwen3.8-Max.
Accuracy (\%); Video-MME(L) is the Long subset of Video-MME.
The embedding variant uses the same text embedding model as Qwen-MM-Plugins (\texttt{qwen3-vl-embedding}); everything else is unchanged.}
\label{tab:search}
\centering
\footnotesize
\setlength{\tabcolsep}{3pt}
\begin{tabular*}{\linewidth}{@{\extracolsep{\fill}}lrrr@{}}
\toprule
Search & LVBench & Video-MME(L) & Video-Holmes \\
\midrule
Embedding retrieval & 86.1 & 85.6 & 72.1 \\
Literal matching (ours) & 86.9 & 86.0 & 71.6 \\
\bottomrule
\end{tabular*}
\end{table}

\section{Prompts}
\label{app:prompts}
\methodname{} uses one system prompt and one user turn per question; everything about individual tools (arguments, usage, return values) is sent as the tool schemas of Table~\ref{tab:tools}.
Prompt~\ref{prompt:system} is the system prompt, and Prompt~\ref{prompt:task} is the user turn that opens each question, with placeholders in angle brackets.
The timestamp sentence follows the video length: videos of at most 60 minutes use \texttt{MM:SS} clock strings instead of \texttt{HH:MM:SS}.
For GPT-5, the system prompt ends with one more sentence, ``Call at most one tool per turn; wait for its result before choosing the next call.'', because its endpoint ignores the request option that disables parallel tool calls.

\refstepcounter{prompt}\label{prompt:system}
\begin{promptbox}{Prompt \theprompt: System prompt}
You are a multimodal agent answering a multiple-choice question about one long video while building and maintaining a reusable memory tree of it. You act only by calling the provided tools, and you finish by calling answer.

You read the video yourself in chronological chunks. read_next attaches the next unread chunk's frames at the streaming fps (READ_FPS); on your next turn you store the complete chunk with write_chunk_memory, and then read on, inspect, or answer. At the very first question of a video the memory is empty and everything is unread, so your only useful first call is read_next; while unread video remains, keep reading if unseen content could change the answer — especially for whole-video questions.

The streaming fps is low, so brief motion, transient on-screen text, fine action order, and exact counts may be missing from memory even when the surrounding events are recorded. Recover those inside the read prefix with view_time on a tight interval at a higher fps, or view_node on a node; what you see is stored only if you call add_observation (after view_time) or refine_node (after view_node of a leaf).

You do NOT receive the memory tree — not even its outline. MEMORY_PROGRESS reports only how far the video is read (read_until) and what remains unread; missing memory after read_until means unread video, not absent content. Everything read so far is stored as segments and retrieved on demand with search_memory. Work from evidence, not from guesses: probe the memory before answering — several literal alternatives in one pattern (synonyms, the option texts, names, on-screen words), a time window when the question pins one down, and different phrasings when the first finds nothing.

Memory holds objective, reusable video facts only — named identities with distinguishing attributes, exact counts, temporal order and state changes, cause-to-consequence pairs, notable dialogue or on-screen text — never the question, the options, your reasoning, or your answer. Segments are chronological, contiguous, and non-overlapping, with meaningful event boundaries; the runtime assigns node IDs, so never invent them.

PRIOR_QUESTIONS lists earlier questions already answered for this same video with the answer you chose; search_memory also returns prior answers whose question mentions the pattern. Use them as cross-question context: keep answers consistent with established narrative understanding. They are NOT objective video facts and must never be written into memory.

Timestamps are global HH:MM:SS clock strings (e.g. "01:11:36"); always write all three fields, never "11:36" or "1:11:36". Inspecting one clip does not imply the question is ready to answer: re-evaluate after every clip and do not guess. Call answer only when the accumulated evidence distinguishes one option.
\end{promptbox}

\refstepcounter{prompt}\label{prompt:task}
\begin{promptbox}{Prompt \theprompt: User turn of each question}
VIDEO_START: 00:00:00
VIDEO_END: <video duration>
TIMESTAMP_FORMAT: HH:MM:SS
READ_FPS: <streaming fps>

PRIOR_QUESTIONS (question -> your answer):
Q1: <earlier question> -> <chosen letter>. <chosen option>
...

MEMORY_PROGRESS:
coverage: read_until <watched until> of <video duration>; unread <watched until> - <video duration>

QUESTION:
<question>

OPTIONS:
A. <option A>
B. <option B>
C. <option C>
D. <option D>
\end{promptbox}

\section{Qualitative Examples}
\label{app:traj}
Figure~\ref{fig:traj_tree} shows the memory tree that Sprout builds on one LVBench video over all 17 of its questions, with Gemini 3.8 Flash and the prompts of Appendix~\ref{app:prompts}; all 17 are answered correctly.
The video is watched only as far as the questions require, and every node below the coarse nodes is written by a revisit that a question triggers.
Figures~\ref{fig:traj_olympics}--\ref{fig:traj_olympics_e} follow nine of these questions: the first five, and the last four, where the questions move past the watched prefix.
The frames are taken from the video at the timestamps shown.
The first question watches the first chunk and records it; every later question starts from the memory left by the questions before it.
Depending on what the memory already holds, a question is answered by one search (Q2), revisits the video and uses the frames as evidence (Q3, Q4, and Q17), revisits and writes a corrected or missing fact back into the tree (Q5, Q14, Q15, and Q16), or first watches further when the question lies past the watched prefix (Q14 and Q17).

\usetikzlibrary{arrows.meta}
\begin{figure}[t]
\centering
\definecolor{q1}{HTML}{4A74B5}
\definecolor{q2}{HTML}{7F8C9A}
\definecolor{q3}{HTML}{9C7BC9}
\definecolor{q4}{HTML}{D08C8C}
\definecolor{q5}{HTML}{D05A5A}
\definecolor{q6}{HTML}{5FA79C}
\definecolor{q7}{HTML}{B39461}
\definecolor{q8}{HTML}{E0913A}
\definecolor{q9}{HTML}{8FA3C9}
\definecolor{q10}{HTML}{8E6BB5}
\definecolor{q11}{HTML}{3E9E86}
\definecolor{q12}{HTML}{8E9A3C}
\definecolor{q13}{HTML}{C46BA0}
\definecolor{q14}{HTML}{5A9E4B}
\definecolor{q15}{HTML}{A0694B}
\definecolor{q16}{HTML}{3AA6B9}
\definecolor{q17}{HTML}{C39A1E}
\begin{tikzpicture}[font=\footnotesize, line cap=round]
\filldraw[fill=gray!6, draw=gray!35, rounded corners=5pt] (-0.25,-1.2) rectangle (11.550,1.62);
\node[anchor=south east, font=\scriptsize\itshape, gray!70!black] at (11.500,-1.18) {Memory};
\node[anchor=west, font=\small] at (-2.3,2.55) {Video};
\node[anchor=west, font=\small] at (-2.3,2.12) {Watch};
\node[anchor=west, font=\scriptsize, gray] at (-2.3,1.82) {low fps};
\node[anchor=west, font=\small] at (-2.3,1.1) {Remember};
\node[anchor=west, font=\scriptsize, gray] at (-2.3,0.8) {coarse node};
\node[anchor=west, font=\small] at (-2.3,0.35) {Revisit};
\node[anchor=west, font=\scriptsize, gray] at (-2.3,0.04999999999999999) {high fps};
\node[anchor=west, font=\small] at (-2.3,-0.62) {Q--A};
\node[anchor=west, font=\scriptsize, gray] at (-2.3,-0.9199999999999999) {record};
\fill[q1!20] (0.015,2.3499999999999996) rectangle (3.945,2.75);
\node[text=q1!80!black, font=\scriptsize\bfseries] at (1.980,2.55) {c1};
\node[gray, font=\tiny, anchor=south] at (0.000,2.75) {00:00};
\fill[q14!20] (3.975,2.3499999999999996) rectangle (6.425,2.75);
\node[text=q14!80!black, font=\scriptsize\bfseries] at (5.200,2.55) {c2};
\node[gray, font=\tiny, anchor=south] at (3.960,2.75) {10:00};
\fill[q14!20] (6.455,2.3499999999999996) rectangle (8.605,2.75);
\node[text=q14!80!black, font=\scriptsize\bfseries] at (7.530,2.55) {c3};
\node[gray, font=\tiny, anchor=south] at (6.440,2.75) {20:00};
\fill[q17!20] (8.635,2.3499999999999996) rectangle (11.085,2.75);
\node[text=q17!80!black, font=\scriptsize\bfseries] at (9.860,2.55) {c4};
\node[gray, font=\tiny, anchor=south] at (8.620,2.75) {30:00};
\node[gray, font=\tiny, anchor=south] at (11.100,2.75) {38:27};
\draw[gray!55, rounded corners=1pt] (0,2.3499999999999996) rectangle (11.100,2.75);
\node[text=q1, font=\footnotesize\bfseries, anchor=west, inner sep=1pt] (w1) at (0.000,2.12) {Q1};
\draw[q1, -{Stealth[length=4pt]}, line width=0.7pt] (w1.east) -- (3.890,2.12);
\draw[q1, line width=0.7pt] (3.930,2.0300000000000002) -- (3.930,2.21);
\node[text=q14, font=\footnotesize\bfseries, anchor=west, inner sep=1pt] (w14) at (3.960,2.12) {Q14};
\draw[q14, -{Stealth[length=4pt]}, line width=0.7pt] (w14.east) -- (8.550,2.12);
\draw[q14, line width=0.7pt] (8.590,2.0300000000000002) -- (8.590,2.21);
\node[text=q17, font=\footnotesize\bfseries, anchor=west, inner sep=1pt] (w17) at (8.620,2.12) {Q17};
\draw[q17, -{Stealth[length=4pt]}, line width=0.7pt] (w17.east) -- (11.030,2.12);
\draw[q17, line width=0.7pt] (11.070,2.0300000000000002) -- (11.070,2.21);
\node[font=\scriptsize\itshape, anchor=north] at (1.980,2.0500000000000003) {stop: Q2--Q13 watch nothing new};
\draw[gray!65] (0,1.1) -- (11.200,1.1) node[right, font=\footnotesize\itshape, black] {$t$};
\node[draw=q1, fill=white, circle, line width=0.8pt, inner sep=0pt, minimum size=8pt] (n1) at (0.310,1.1) {};\fill[q1] (0.310,1.1) circle (1.3pt);
\node[font=\tiny, gray!70!black, anchor=south, inner sep=1.5pt] at (0.310,1.23) {n1};
\draw[q8, line width=0.6pt] (0.310,0.9600000000000001) -- (0.310,0.33999999999999997);
\node[draw=q8, fill=q8!25, circle, line width=0.7pt, inner sep=0pt, minimum size=6.5pt] at (0.310,0.25) {};
\node[draw=q1, fill=white, circle, line width=0.8pt, inner sep=0pt, minimum size=8pt] (n2) at (1.090,1.1) {};\fill[q1] (1.090,1.1) circle (1.3pt);
\node[font=\tiny, gray!70!black, anchor=south, inner sep=1.5pt] at (1.090,1.23) {n2};
\draw[gray!70, line width=0.6pt] (1.090,0.9600000000000001) -- (1.090,0.6800000000000002);\draw[q8, line width=0.6pt] (1.090,0.6800000000000002) -- (0.880,0.6800000000000002) -- (0.880,0.33999999999999997);
\draw[gray!70, line width=0.6pt] (1.090,0.9600000000000001) -- (1.090,0.6800000000000002);\draw[q5, line width=0.6pt] (1.090,0.6800000000000002) -- (1.300,0.6800000000000002) -- (1.300,0.33999999999999997);
\node[draw=q8, fill=q8!25, circle, line width=0.7pt, inner sep=0pt, minimum size=6.5pt] at (0.880,0.25) {};
\node[draw=q5, fill=q5!25, circle, line width=0.7pt, inner sep=0pt, minimum size=6.5pt] at (1.300,0.25) {};
\node[draw=q1, fill=white, circle, line width=0.8pt, inner sep=0pt, minimum size=8pt] (n3) at (2.450,1.1) {};\fill[q1] (2.450,1.1) circle (1.3pt);
\node[font=\tiny, gray!70!black, anchor=south, inner sep=1.5pt] at (2.450,1.23) {n3};
\draw[gray!70, line width=0.6pt] (2.450,0.9600000000000001) -- (2.450,0.6800000000000002);\draw[q10, line width=0.6pt] (2.450,0.6800000000000002) -- (1.820,0.6800000000000002) -- (1.820,0.33999999999999997);
\draw[gray!70, line width=0.6pt] (2.450,0.9600000000000001) -- (2.450,0.6800000000000002);\draw[q11, line width=0.6pt] (2.450,0.6800000000000002) -- (2.240,0.6800000000000002) -- (2.240,0.33999999999999997);
\draw[gray!70, line width=0.6pt] (2.450,0.9600000000000001) -- (2.450,0.6800000000000002);\draw[q12, line width=0.6pt] (2.450,0.6800000000000002) -- (2.660,0.6800000000000002) -- (2.660,0.33999999999999997);
\draw[gray!70, line width=0.6pt] (2.450,0.9600000000000001) -- (2.450,0.6800000000000002);\draw[q13, line width=0.6pt] (2.450,0.6800000000000002) -- (3.080,0.6800000000000002) -- (3.080,0.33999999999999997);
\node[draw=q10, fill=q10!25, circle, line width=0.7pt, inner sep=0pt, minimum size=6.5pt] at (1.820,0.25) {};
\node[draw=q11, fill=q11!25, circle, line width=0.7pt, inner sep=0pt, minimum size=6.5pt] at (2.240,0.25) {};
\node[draw=q12, fill=q12!25, circle, line width=0.7pt, inner sep=0pt, minimum size=6.5pt] at (2.660,0.25) {};
\node[draw=q13, fill=q13!25, circle, line width=0.7pt, inner sep=0pt, minimum size=6.5pt] at (3.080,0.25) {};
\node[draw=q1, fill=white, circle, line width=0.8pt, inner sep=0pt, minimum size=8pt] (n4) at (3.650,1.1) {};\fill[q1] (3.650,1.1) circle (1.3pt);
\node[font=\tiny, gray!70!black, anchor=south, inner sep=1.5pt] at (3.650,1.23) {n4};
\draw[q13, line width=0.6pt] (3.650,0.9600000000000001) -- (3.650,0.33999999999999997);
\node[draw=q13, fill=q13!25, circle, line width=0.7pt, inner sep=0pt, minimum size=6.5pt] at (3.650,0.25) {};
\node[draw=q14, fill=white, circle, line width=0.8pt, inner sep=0pt, minimum size=8pt] (n5) at (4.270,1.1) {};\fill[q14] (4.270,1.1) circle (1.3pt);
\node[font=\tiny, gray!70!black, anchor=south, inner sep=1.5pt] at (4.270,1.23) {n5};
\node[draw=q14, fill=white, circle, line width=0.8pt, inner sep=0pt, minimum size=8pt] (n6) at (4.890,1.1) {};\fill[q14] (4.890,1.1) circle (1.3pt);
\node[font=\tiny, gray!70!black, anchor=south, inner sep=1.5pt] at (4.890,1.23) {n6};
\node[draw=q14, fill=white, circle, line width=0.8pt, inner sep=0pt, minimum size=8pt] (n7) at (5.510,1.1) {};\fill[q14] (5.510,1.1) circle (1.3pt);
\node[font=\tiny, gray!70!black, anchor=south, inner sep=1.5pt] at (5.510,1.23) {n7};
\node[draw=q14, fill=white, circle, line width=0.8pt, inner sep=0pt, minimum size=8pt] (n8) at (6.130,1.1) {};\fill[q14] (6.130,1.1) circle (1.3pt);
\node[font=\tiny, gray!70!black, anchor=south, inner sep=1.5pt] at (6.130,1.23) {n8};
\node[draw=q14, fill=white, circle, line width=0.8pt, inner sep=0pt, minimum size=8pt] (n9) at (6.910,1.1) {};\fill[q14] (6.910,1.1) circle (1.3pt);
\node[font=\tiny, gray!70!black, anchor=south, inner sep=1.5pt] at (6.910,1.23) {n9};
\draw[gray!70, line width=0.6pt] (6.910,0.9600000000000001) -- (6.910,0.6800000000000002);\draw[q14, line width=0.6pt] (6.910,0.6800000000000002) -- (6.700,0.6800000000000002) -- (6.700,0.33999999999999997);
\draw[gray!70, line width=0.6pt] (6.910,0.9600000000000001) -- (6.910,0.6800000000000002);\draw[q15, line width=0.6pt] (6.910,0.6800000000000002) -- (7.120,0.6800000000000002) -- (7.120,0.33999999999999997);
\node[draw=q14, fill=q14!25, circle, line width=0.7pt, inner sep=0pt, minimum size=6.5pt] at (6.700,0.25) {};
\node[draw=q15, fill=q15!25, circle, line width=0.7pt, inner sep=0pt, minimum size=6.5pt] at (7.120,0.25) {};
\node[draw=q14, fill=white, circle, line width=0.8pt, inner sep=0pt, minimum size=8pt] (n10) at (7.690,1.1) {};\fill[q14] (7.690,1.1) circle (1.3pt);
\node[font=\tiny, gray!70!black, anchor=south, inner sep=1.5pt] at (7.690,1.23) {n10};
\draw[q16, line width=0.6pt] (7.690,0.9600000000000001) -- (7.690,0.33999999999999997);
\node[draw=q16, fill=q16!25, circle, line width=0.7pt, inner sep=0pt, minimum size=6.5pt] at (7.690,0.25) {};
\node[draw=q14, fill=white, circle, line width=0.8pt, inner sep=0pt, minimum size=8pt] (n11) at (8.310,1.1) {};\fill[q14] (8.310,1.1) circle (1.3pt);
\node[font=\tiny, gray!70!black, anchor=south, inner sep=1.5pt] at (8.310,1.23) {n11};
\node[draw=q17, fill=white, circle, line width=0.8pt, inner sep=0pt, minimum size=8pt] (n12) at (8.930,1.1) {};\fill[q17] (8.930,1.1) circle (1.3pt);
\node[font=\tiny, gray!70!black, anchor=south, inner sep=1.5pt] at (8.930,1.23) {n12};
\node[draw=q17, fill=white, circle, line width=0.8pt, inner sep=0pt, minimum size=8pt] (n13) at (9.550,1.1) {};\fill[q17] (9.550,1.1) circle (1.3pt);
\node[font=\tiny, gray!70!black, anchor=south, inner sep=1.5pt] at (9.550,1.23) {n13};
\node[draw=q17, fill=white, circle, line width=0.8pt, inner sep=0pt, minimum size=8pt] (n14) at (10.170,1.1) {};\fill[q17] (10.170,1.1) circle (1.3pt);
\node[font=\tiny, gray!70!black, anchor=south, inner sep=1.5pt] at (10.170,1.23) {n14};
\node[draw=q17, fill=white, circle, line width=0.8pt, inner sep=0pt, minimum size=8pt] (n15) at (10.790,1.1) {};\fill[q17] (10.790,1.1) circle (1.3pt);
\node[font=\tiny, gray!70!black, anchor=south, inner sep=1.5pt] at (10.790,1.23) {n15};
\node[draw=q1, fill=q1!14, rounded corners=2pt, line width=0.6pt, font=\scriptsize\bfseries, text=q1!85!black, minimum width=0.563cm, minimum height=0.4cm, inner sep=0pt] at (0.326,-0.62) {Q1};
\node[draw=q2, fill=q2!14, rounded corners=2pt, line width=0.6pt, font=\scriptsize\bfseries, text=q2!85!black, minimum width=0.563cm, minimum height=0.4cm, inner sep=0pt] at (0.979,-0.62) {Q2};
\node[draw=q3, fill=q3!14, rounded corners=2pt, line width=0.6pt, font=\scriptsize\bfseries, text=q3!85!black, minimum width=0.563cm, minimum height=0.4cm, inner sep=0pt] at (1.632,-0.62) {Q3};
\node[draw=q4, fill=q4!14, rounded corners=2pt, line width=0.6pt, font=\scriptsize\bfseries, text=q4!85!black, minimum width=0.563cm, minimum height=0.4cm, inner sep=0pt] at (2.285,-0.62) {Q4};
\node[draw=q5, fill=q5!14, rounded corners=2pt, line width=0.6pt, font=\scriptsize\bfseries, text=q5!85!black, minimum width=0.563cm, minimum height=0.4cm, inner sep=0pt] at (2.938,-0.62) {Q5};
\node[draw=q6, fill=q6!14, rounded corners=2pt, line width=0.6pt, font=\scriptsize\bfseries, text=q6!85!black, minimum width=0.563cm, minimum height=0.4cm, inner sep=0pt] at (3.591,-0.62) {Q6};
\node[draw=q7, fill=q7!14, rounded corners=2pt, line width=0.6pt, font=\scriptsize\bfseries, text=q7!85!black, minimum width=0.563cm, minimum height=0.4cm, inner sep=0pt] at (4.244,-0.62) {Q7};
\node[draw=q8, fill=q8!14, rounded corners=2pt, line width=0.6pt, font=\scriptsize\bfseries, text=q8!85!black, minimum width=0.563cm, minimum height=0.4cm, inner sep=0pt] at (4.897,-0.62) {Q8};
\node[draw=q9, fill=q9!14, rounded corners=2pt, line width=0.6pt, font=\scriptsize\bfseries, text=q9!85!black, minimum width=0.563cm, minimum height=0.4cm, inner sep=0pt] at (5.550,-0.62) {Q9};
\node[draw=q10, fill=q10!14, rounded corners=2pt, line width=0.6pt, font=\scriptsize\bfseries, text=q10!85!black, minimum width=0.563cm, minimum height=0.4cm, inner sep=0pt] at (6.203,-0.62) {Q10};
\node[draw=q11, fill=q11!14, rounded corners=2pt, line width=0.6pt, font=\scriptsize\bfseries, text=q11!85!black, minimum width=0.563cm, minimum height=0.4cm, inner sep=0pt] at (6.856,-0.62) {Q11};
\node[draw=q12, fill=q12!14, rounded corners=2pt, line width=0.6pt, font=\scriptsize\bfseries, text=q12!85!black, minimum width=0.563cm, minimum height=0.4cm, inner sep=0pt] at (7.509,-0.62) {Q12};
\node[draw=q13, fill=q13!14, rounded corners=2pt, line width=0.6pt, font=\scriptsize\bfseries, text=q13!85!black, minimum width=0.563cm, minimum height=0.4cm, inner sep=0pt] at (8.162,-0.62) {Q13};
\node[draw=q14, fill=q14!14, rounded corners=2pt, line width=0.6pt, font=\scriptsize\bfseries, text=q14!85!black, minimum width=0.563cm, minimum height=0.4cm, inner sep=0pt] at (8.815,-0.62) {Q14};
\node[draw=q15, fill=q15!14, rounded corners=2pt, line width=0.6pt, font=\scriptsize\bfseries, text=q15!85!black, minimum width=0.563cm, minimum height=0.4cm, inner sep=0pt] at (9.468,-0.62) {Q15};
\node[draw=q16, fill=q16!14, rounded corners=2pt, line width=0.6pt, font=\scriptsize\bfseries, text=q16!85!black, minimum width=0.563cm, minimum height=0.4cm, inner sep=0pt] at (10.121,-0.62) {Q16};
\node[draw=q17, fill=q17!14, rounded corners=2pt, line width=0.6pt, font=\scriptsize\bfseries, text=q17!85!black, minimum width=0.563cm, minimum height=0.4cm, inner sep=0pt] at (10.774,-0.62) {Q17};
\end{tikzpicture}
\caption{\textbf{The memory tree that Sprout builds on one LVBench video} (Gemini 3.8 Flash, the video of Figures~\ref{fig:traj_olympics} and~\ref{fig:traj_olympics_b}), after all 17 of its questions, all answered correctly. Each question has its own color. Q1 watches the first 10-minute chunk and records it as n1--n4; Q2--Q13 watch nothing further, Q14 watches the next two chunks, and Q17 the rest. Each child below a coarse node is written by a revisit and carries the color of the question that wrote it; Q2, Q3, Q4, Q6, Q7, and Q9 answer without writing anything. The child of n2 written by Q5 is the one shown in Figure~\ref{fig:traj_olympics_b}.}
\label{fig:traj_tree}
\end{figure}
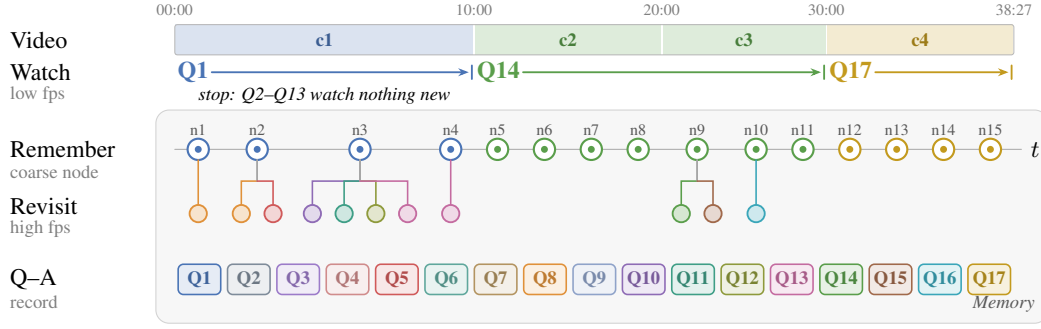

\newcommand{\trajq}[4]{%
  \par\noindent\colorbox{gray!10}{\parbox{\dimexpr\linewidth-2\fboxsep}{\raggedright
    \act{actAnswer!85!black}{#1}\enspace\textbf{#2}\hfill\textcolor{actAnswer}{\scriptsize #4}\\[1pt]
    \hspace*{2.2em}#3}}\par\vspace{4pt}}
\newcommand{\trajrow}[2]{%
  \par\noindent\begin{minipage}[t]{0.145\linewidth}#1\end{minipage}\hfill
  \begin{minipage}[t]{0.85\linewidth}\raggedright #2\end{minipage}\par\vspace{5pt}}
\newcommand{\trajlabel}[3]{\makebox[1.1em][l]{\textcolor{actAnswer}{\scriptsize\bfseries #1}}\act{#2}{#3}}
\newcommand{\trajr}[5]{\trajrow{\trajlabel{#1}{#2}{#3}}{\call{#4}\begin{res}{#2}#5\end{res}}}
\newcommand{\traja}[3]{\trajrow{\trajlabel{#1}{actAnswer}{Answer}}{\call{#2}\hfill\textcolor{actAnswer}{\scriptsize #3}}}
\newcommand{\trajbox}[2]{%
  \begin{tcolorbox}[enhanced, colback=white, colframe=actAnswer, boxrule=0.5pt, arc=2pt, left=6pt, right=6pt, top=4pt, bottom=0pt,
    title={#1}, fonttitle=\small\bfseries]
  \footnotesize #2
  \end{tcolorbox}}

\begin{figure}[p]
\centering
\trajbox{LVBench \texttt{Z86xysw5Ncc} (38:27), questions 1--3 of the video}{%
\trajq{Q1}{What category does this video belong to?}
  {\gold{A.~Olympic games}\quad B.~The BBC News\quad C.~The NBA\quad D.~The country life vlog}{3 rounds\quad 11.7K tokens}
\trajr{r1}{actWatch}{Watch}{read\_next()}
  {Chunk 00:00--10:00 attached at 0.1\,fps (60 frames).\par\vspace{2pt}
   \trajframe{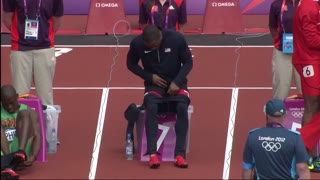}{00:40}\hfill\trajframe{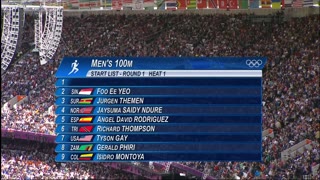}{01:40}\hfill\trajframe{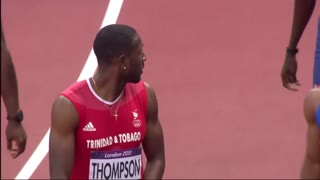}{04:20}\hfill\trajframe{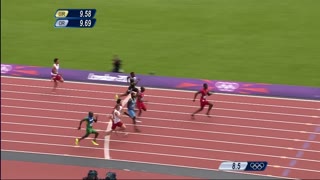}{07:30}\hfill\trajframe{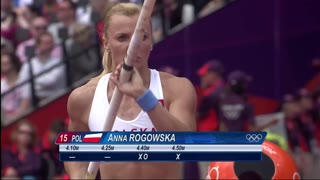}{08:50}}
\trajr{r2}{actRemember}{Remember}{write\_chunk\_memory(4 segments)}
  {Stored node\_1--node\_4; the frames are released.
   \begin{memcard}
   \textbf{node\_1}\quad 00:00--01:40\quad Introduction and start list for Men's 100m Round 1 Heat 1 at London 2012 Olympic Games.\\
   \textbf{node\_2}\quad 01:40--05:20\quad Men's 100m Round 1 Heat 1 race won by Tyson Gay.\\
   \textbf{node\_3}\quad 05:20--08:40\quad Men's 100m Round 1 Heat 2 race won by Justin Gatlin.\\
   \textbf{node\_4}\quad 08:40--10:00\quad Women's pole vault competition at London 2012 Olympic stadium.
   \end{memcard}}
\traja{r3}{answer("A")}{}
\vspace{4pt}
\trajq{Q2}{How many meters is the race about men in the video?}
  {A.~1500\quad B.~3000\quad \gold{C.~100}\quad D.~50}{2 rounds\quad 5.0K tokens}
\trajr{r1}{actRevisit}{Revisit}{search\_memory("race|meter|1500|3000|100|50|track|run")}
  {5 facts in 4 nodes, led by node\_1: ``Introduction and start list for Men's 100m Round 1 Heat 1 \ldots''}
\traja{r2}{answer("C")}{answered from memory alone, no frames}
\vspace{4pt}
\trajq{Q3}{How many athletes in Men's 100M Start-List-Round 1 Heat 1?}
  {A.~9\quad B.~7\quad \gold{C.~8}\quad D.~10}{4 rounds\quad 12.7K tokens}
\trajr{r1}{actRevisit}{Revisit}{search\_memory("Heat 1|Start-List|Round 1|Heat")}
  {8 facts in 3 nodes, including @01:40 ``Start list for Men's 100m Round 1 Heat 1 is displayed on screen.'' The count is not recorded.}
\trajr{r2}{actRevisit}{Revisit}{view\_time(01:30, 02:00, fps=1)}
  {The end is clamped to 01:40, the end of node\_1.}
\trajr{r3}{actRevisit}{Revisit}{view\_time(01:40, 01:45, fps=1)}
  {5 frames of node\_2.\par\vspace{2pt}
   \trajframe[0.25]{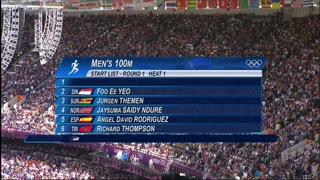}{01:38}\hspace{0.02\linewidth}\trajframe[0.25]{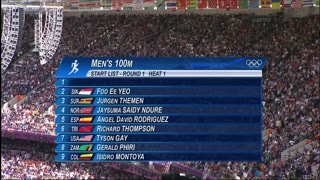}{01:41}}
\traja{r4}{answer("C")}{frames used as evidence only, nothing written}
}
\vspace{-0.6em}
\caption{\textbf{The first three questions on one LVBench video} (\methodname{} with Gemini 3.8 Flash; Figure~\ref{fig:traj_olympics_b} continues with questions 4 and 5).
Each round shows the tool call (\,$\triangleright$\,) and, in the shaded block, its result.
Q1 watches the first chunk and records it as four nodes; no later question in this figure watches further.
Q2 is answered by one search.
Q3 finds where the start list is shown but not its length, so it revisits that interval at 1\,fps, first clamped at the boundary of node\_1, and counts the athletes on the frames.
Tool calls keep only their key arguments and tool results are abridged; memory text is verbatim from the request log, and token counts are total input tokens per question.}
\label{fig:traj_olympics}
\end{figure}

\begin{figure}[p]
\centering
\trajbox{LVBench \texttt{Z86xysw5Ncc} (38:27), questions 4--5 of the video}{%
\trajq{Q4}{What's the number on the back of Tyson Gay?}
  {A.~3218\quad B.~3712\quad C.~3718\quad \gold{D.~3217}}{4 rounds\quad 26.4K tokens}
\trajr{r1}{actRevisit}{Revisit}{search\_memory("Gay|Tyson|3218|3712|3718|3217")}
  {2 facts in node\_2; no number.}
\trajr{r2}{actRevisit}{Revisit}{search\_memory(start=01:40, end=05:20)}
  {All 4 facts of node\_2; no number.}
\trajr{r3}{actRevisit}{Revisit}{view\_node(node\_2, fps=1, focus="number on the back of Tyson Gay's uniform or bib")}
  {The whole node 01:40--05:20 attached at 1\,fps.\par\vspace{2pt}
   \trajframe[0.25]{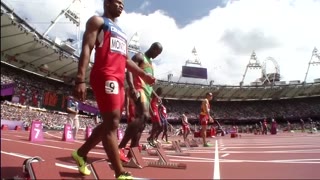}{03:19}\hspace{0.02\linewidth}\trajframe[0.25]{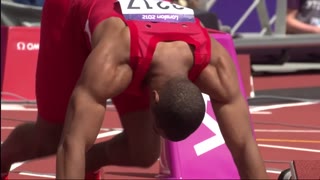}{03:31}}
\traja{r4}{answer("D")}{frames used as evidence only, nothing written}
\vspace{4pt}
\trajq{Q5}{When does the first heat of the men's 100m startstart of a race?}
  {A.~The video time of 02:56\quad B.~The video time of 03:46\\\hspace*{2.2em}\gold{C.~The video time of 03:56}\quad D.~The video time of 02:46}{4 rounds\quad 13.1K tokens}
\trajr{r1}{actRevisit}{Revisit}{search\_memory("heat 1 | start | race | gun | 100m", start=02:00, end=04:30)}
  {@03:50 ``The starter fires the gun and Heat 1 commences.'', and the prior answers of Q2 and Q3.}
\trajr{r2}{actRevisit}{Revisit}{view\_time(03:40, 04:05, fps=1)}
  {25 frames of node\_2.\par\vspace{2pt}
   \trajframe[0.25]{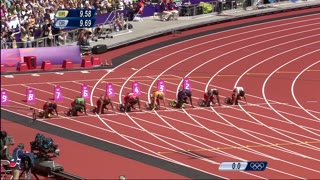}{03:53}\hspace{0.02\linewidth}\trajframe[0.25]{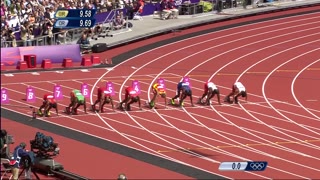}{03:56}}
\trajr{r3}{actRevisit}{Revisit}{add\_observation(summary, 3 facts)}
  {Stored as child node\_2\_1; the frames are released.
   \begin{memcard}
   \textbf{node\_2\_1}\quad 03:40--04:05\quad Men's 100m Round 1 Heat 1 starts at 03:56 with runners leaping from the blocks.\\
   @03:56\quad The race of Men's 100m Round 1 Heat 1 officially starts as the gun fires and running clock starts (reading 0.1 at 03:56).
   \end{memcard}}
\traja{r4}{answer("C")}{}
}
\vspace{-0.6em}
\caption{\textbf{Questions 4 and 5 on the video of Figure~\ref{fig:traj_olympics}.}
Q4 finds that the bib number is not in memory, by pattern and then by time window, and revisits the whole node\_2 at 1\,fps to read it.
Q5 revisits the start of the race, finds the gun at 03:56 rather than the 03:50 recorded at 0.1\,fps, and writes the corrected time back as the child node\_2\_1.}
\label{fig:traj_olympics_b}
\end{figure}

\begin{figure}[p]
\centering
\trajbox{LVBench \texttt{Z86xysw5Ncc} (38:27), question 14 of the video}{%
\trajq{Q14}{When does Men's 100M Round 1 Heat 4 start preparing?}
  {A.~At the video time of 19:38\quad B.~At the video time of 20:38\\\hspace*{2.2em}C.~At the video time of 19:28\quad \gold{D.~At the video time of 20:28}}{7 rounds\quad 46.8K tokens}
\trajr{r1}{actWatch}{Watch}{read\_next()}
  {Chunk 10:00--20:00 attached at 0.1\,fps (60 frames).\par\vspace{2pt}
   \trajframe{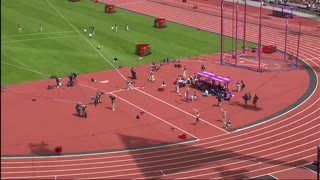}{10:20}\hfill\trajframe{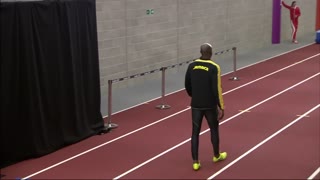}{12:30}\hfill\trajframe{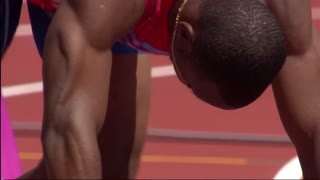}{15:00}\hfill\trajframe{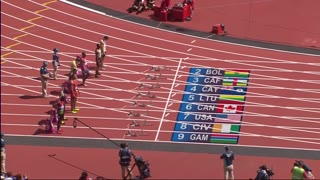}{18:00}}
\trajr{r2}{actRemember}{Remember}{write\_chunk\_memory(4 segments)}
  {Stored node\_5--node\_8; the frames are released.
   \begin{memcard}
   \textbf{node\_5}\quad 10:00--11:25\quad Field event highlights including pole vault and javelin throw, followed by the Heat 2 result graphic.\\
   \textbf{node\_6}\quad 11:25--13:15\quad Athletes warming up in the indoor facility and gathering in the call room.\\
   \textbf{node\_7}\quad 13:15--16:45\quad Men's 100M Round 1 Heat 3 intros, race, and celebration won by Ryan Bailey.\\
   \textbf{node\_8}\quad 16:45--20:00\quad Repeated coverage of Men's 100M Round 1 Heat 3 including start list, race, and slow-motion replays.
   \end{memcard}}
\trajr{r3}{actWatch}{Watch}{read\_next()}
  {Chunk 20:00--30:00 attached at 0.1\,fps (60 frames).\par\vspace{2pt}
   \trajframe{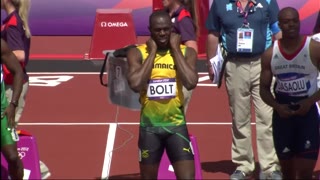}{21:00}\hfill\trajframe{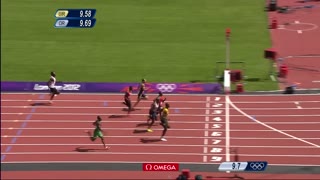}{23:30}\hfill\trajframe{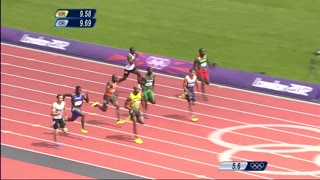}{26:40}\hfill\trajframe{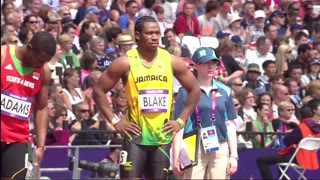}{29:00}}
\trajr{r4}{actRemember}{Remember}{write\_chunk\_memory(3 segments)}
  {Stored node\_9--node\_11; the frames are released.
   \begin{memcard}
   \textbf{node\_9}\quad 20:00--24:45\quad Men's 100M Round 1 Heat 4 start list, athlete introductions, race won by Usain Bolt in 10.08s, and replays.\\
   \textbf{node\_10}\quad 24:45--28:10\quad Men's 100M Round 1 Heat 5 start list, intros, race won by Asafa Powell in 10.04s, and replays.\\
   \textbf{node\_11}\quad 28:10--30:00\quad Men's 100M Round 1 Heat 6 introductions and lineup at the starting blocks.
   \end{memcard}}
\trajr{r5}{actRevisit}{Revisit}{view\_time(20:20, 20:45, fps=1, focus="20:28 vs 20:38")}
  {25 frames of node\_9.\par\vspace{2pt}
   \trajframe[0.25]{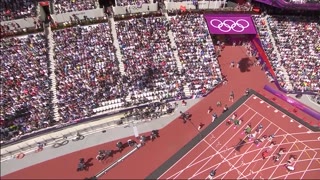}{20:28}\hspace{0.02\linewidth}\trajframe[0.25]{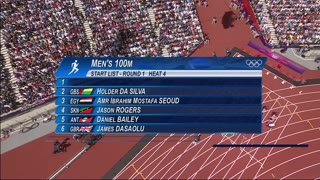}{20:38}}
\trajr{r6}{actRevisit}{Revisit}{add\_observation(summary, 3 facts)}
  {Stored as child node\_9\_1; the frames are released.
   \begin{memcard}
   \textbf{node\_9\_1}\quad 20:20--20:45\quad Camera settles on the track for Heat 4 and the Start List - Round 1 Heat 4 graphic appears.\\
   @20:28\quad The camera pans down over the crowd to focus on the 100m start line where athletes are milling about and preparing.
   \end{memcard}}
\traja{r7}{answer("D")}{}
}
\vspace{-0.6em}
\caption{\textbf{Question 14 on the video of Figure~\ref{fig:traj_olympics}.}
Q2--Q13 are all answered within the first chunk.
Q14 asks about a moment after 10:00, so it watches and records the next two chunks, then revisits the start of Heat 4 at 1\,fps to tell 20:28 from 20:38 and writes what it sees as node\_9\_1.}
\label{fig:traj_olympics_c}
\end{figure}

\begin{figure}[p]
\centering
\trajbox{LVBench \texttt{Z86xysw5Ncc} (38:27), questions 15--16 of the video}{%
\trajq{Q15}{What is the best record for Men's 100M Round 1 Heat 4?}
  {A.~10.23\quad B.~9.58\quad C.~9.69\quad \gold{D.~10.09}}{4 rounds\quad 16.5K tokens}
\trajr{r1}{actRevisit}{Revisit}{search\_memory("Heat 4|Round 1 Heat 4|10.|9.", start=20:00)}
  {5 facts in 3 nodes, including @23:30 ``The Heat 4 race starts, and Usain Bolt pulls ahead to win comfortably in 10.08s.'' 10.08 is not an option.}
\trajr{r2}{actRevisit}{Revisit}{view\_time(23:35, 24:20, fps=1, focus="winning time")}
  {45 frames of node\_9.\par\vspace{2pt}
   \trajframe[0.25]{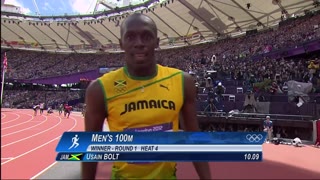}{23:52}}
\trajr{r3}{actRevisit}{Revisit}{add\_observation(summary, 1 fact)}
  {Stored as child node\_9\_2; the frames are released.
   \begin{memcard}
   \textbf{node\_9\_2}\quad 23:35--24:20\quad Usain Bolt wins Men's 100M Round 1 Heat 4 with an official time of 10.09s.
   \end{memcard}}
\traja{r4}{answer("D")}{}
\vspace{4pt}
\trajq{Q16}{Who is the first in Men's 100M Roound 1 Heat 5?}
  {A.~Su BingTian\quad B.~Usain Bolt\quad \gold{C.~Asafa Powell}\quad D.~Tyson Gay}{4 rounds\quad 17.2K tokens}
\trajr{r1}{actRevisit}{Revisit}{search\_memory("Heat 5|Heat 4|Bolt|Bingtian|Powell")}
  {18 facts in 7 nodes, including @26:40 ``Heat 5 race is run and Asafa Powell finishes first in 10.04s.''}
\trajr{r2}{actRevisit}{Revisit}{view\_time(26:40, 27:05, fps=1, focus="who finishes first")}
  {25 frames of node\_10.\par\vspace{2pt}
   \trajframe[0.25]{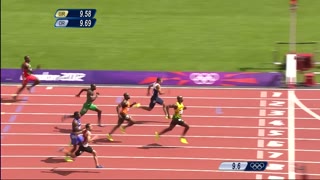}{26:44}\hspace{0.02\linewidth}\trajframe[0.25]{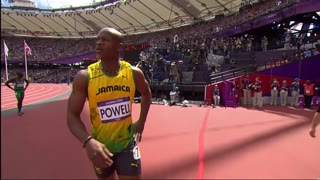}{27:04}}
\trajr{r3}{actRevisit}{Revisit}{add\_observation(summary, 2 facts)}
  {Stored as child node\_10\_1; the frames are released.
   \begin{memcard}
   \textbf{node\_10\_1}\quad 26:40--27:05\quad Finish of Men's 100M Round 1 Heat 5 won by Asafa Powell with a time of 10.04s.
   \end{memcard}}
\traja{r4}{answer("C")}{}
}
\vspace{-0.6em}
\caption{\textbf{Questions 15 and 16 on the video of Figure~\ref{fig:traj_olympics}.}
Q15 finds a winning time of 10.08s recorded at 0.1\,fps, which is not among the options; the revisit reads 10.09s from the result graphic and writes it back as node\_9\_2.
Q16 finds the winner already in memory and revisits the finish to confirm it before answering.}
\label{fig:traj_olympics_d}
\end{figure}

\begin{figure}[p]
\centering
\trajbox{LVBench \texttt{Z86xysw5Ncc} (38:27), question 17 of the video}{%
\trajq{Q17}{What place is Su Bingtian in for Men's 100M Round 1 Heat 6?}
  {A.~Fourth\quad \gold{B.~Third}\quad C.~First\quad D.~Second}{8 rounds\quad 45.2K tokens}
\trajr{r1}{actRevisit}{Revisit}{search\_memory("Su|Bingtian|Heat 6|Heat6|China")}
  {6 facts in 5 nodes; for Heat 6 only node\_11, ``introductions and lineup at the starting blocks'', which ends at 30:00, where the read prefix ends.}
\trajr{r2}{actWatch}{Watch}{read\_next()}
  {Chunk 30:00--38:27 attached at 0.1\,fps (51 frames).\par\vspace{2pt}
   \trajframe{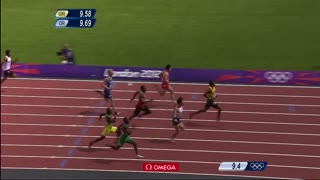}{30:40}\hfill\trajframe{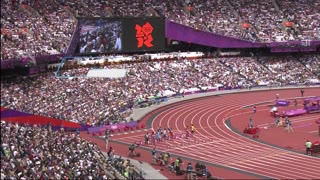}{33:00}\hfill\trajframe{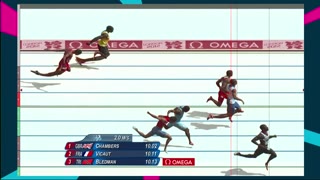}{36:00}\hfill\trajframe{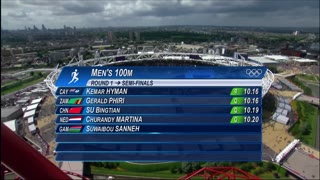}{38:10}}
\trajr{r3}{actRemember}{Remember}{write\_chunk\_memory(4 segments)}
  {Stored node\_12--node\_15; nothing is left unread.
   \begin{memcard}
   \textbf{node\_12}\quad 30:00--31:50\quad Men's 100M Round 1 Heat 6 is run and won by Yohan Blake, with Ryota Yamagata second and Su Bingtian third.\\
   \hspace*{1em}@31:00\quad The stadium big screen shows results for Heat 6: 1st Yohan Blake (10.00 Q), 2nd Ryota Yamagata (10.07 Q PB), 3rd Su Bingtian (10.19 Q).\\
   \textbf{node\_13}\quad 31:50--34:30\quad Preparation and athlete introductions for Men's 100M Round 1 Heat 7.\\
   \textbf{node\_14}\quad 34:30--37:35\quad Men's 100M Round 1 Heat 7 is run and won by Dwain Chambers.\\
   \textbf{node\_15}\quad 37:35--38:27\quad Semi-final qualifiers list is displayed, and the broadcast concludes with stadium aerial views.
   \end{memcard}}
\trajr{r4}{actRevisit}{Revisit}{view\_time(30:55, 31:15, fps=1, focus="Su Bingtian's place")}
  {20 frames of node\_12.}
\trajr{r5}{actRevisit}{Revisit}{view\_time(31:30, 31:45, fps=1, focus="official results graphic")}
  {15 frames of node\_12.}
\trajr{r6}{actRevisit}{Revisit}{view\_time(31:00, 31:08, fps=2, focus="places 1, 2, 3")}
  {16 frames of node\_12.\par\vspace{2pt}
   \trajframe[0.25]{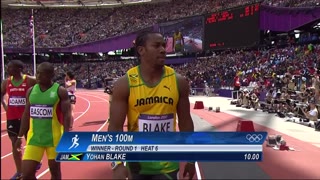}{31:00}\hspace{0.02\linewidth}\trajframe[0.25]{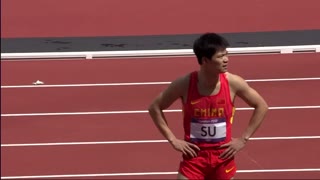}{31:08}}
\trajr{r7}{actRevisit}{Revisit}{view\_time(38:05, 38:15, fps=1, focus="qualifiers graphic")}
  {10 frames of node\_15.}
\traja{r8}{answer("B")}{frames used as evidence only, nothing written}
}
\vspace{-0.6em}
\caption{\textbf{Question 17 on the video of Figure~\ref{fig:traj_olympics}.}
Q17 asks about Heat 6, whose race lies past the watched prefix, so it watches and records the last chunk; the new memory already names Su Bingtian third, and four revisits of the result graphics confirm it without writing anything.}
\label{fig:traj_olympics_e}
\end{figure}

\end{document}